\documentclass[pdflatex,bst/sn-mathphys-num]{sn-jnl}

\usepackage{graphicx}%
\usepackage{multirow}%
\usepackage{amsmath,amssymb,amsfonts}%
\usepackage{amsthm}%
\usepackage{mathrsfs}%
\usepackage[title]{appendix}%
\usepackage{xcolor}%
\usepackage{textcomp}%
\usepackage{manyfoot}%
\usepackage{booktabs}%
\usepackage{algorithm}
\usepackage{float} 
\usepackage{subcaption} 
\usepackage{algpseudocode}%
\usepackage{listings}%
\usepackage{pifont}%
\usepackage{cite}
\usepackage{graphicx}
\usepackage{boldline}
\usepackage[table]{xcolor}
\usepackage{pifont}
\usepackage{tabularray}
\usepackage{hyperref}

\theoremstyle{thmstyleone}%
\theoremstyle{thmstyletwo}%

\theoremstyle{thmstylethree}%

\begin{document}
\title[Article Title]{Learning from Single Timestamps: Complexity Estimation in Laparoscopic Cholecystectomy}


\author*[1,3]{\fnm{Dimitrios} \sur{Anastasiou}}\email{dimitrios.anastasiou.21@ucl.ac.uk}

\author[2]{\fnm{Santiago} \sur{Barbarisi}}\email{santiago.barbarisi@medtronic.com}

\author[2]{\fnm{Lucy} \sur{Culshaw}}\email{lucy.h.culshaw@medtronic.com}

\author[2]{\fnm{Jayna} \sur{Patel}}\email{jayna.patel2@medtronic.com}

\author[1,3]{\fnm{Evangelos B.} \sur{Mazomenos}}\email{e.mazomenos@ucl.ac.uk}
 
\author[2]{\fnm{Imanol} \sur{Luengo}}\email{imanol.luengo@medtronic.com}

\author[1,2,4]{\fnm{Danail} \sur{Stoyanov}}\email{danail.stoyanov@ucl.ac.uk}
 
\affil[1]{\orgname{UCL Hawkes Institute, University College London}, \orgaddress{\city{London}, \country{UK}}}

\affil[2]{\orgname{Medtronic plc.}, \orgaddress{\city{London}, \country{UK}}}

\affil[3]{\orgname{Dept of Medical Physics \& Biomedical Engineering, University College London}, \orgaddress{\city{London}, \country{UK}}}

\affil[4]{\orgname{Dept of Computer Science, University College London}, \orgaddress{\city{London}, \country{UK}}}

\abstract{\textbf{Purpose:} Accurate assessment of surgical complexity is essential in Laparoscopic Cholecystectomy (LC), where severe inflammation is associated with longer operative times and increased risk of postoperative complications. The Parkland Grading Scale (PGS) provides a clinically validated framework for stratifying inflammation severity; however, its automation in surgical videos remains largely unexplored, particularly in realistic scenarios where complete videos must be analyzed without prior manual curation.\\
\textbf{Methods:} In this work, we introduce STC-Net, a novel framework for Single-Timestamp-based Complexity estimation in LC via the PGS, designed to operate under weak temporal supervision. Unlike prior methods limited to static images or manually trimmed clips, STC-Net operates directly on full videos. It jointly performs temporal localization and grading through a localization, window proposal, and grading module. We introduce a novel loss formulation combining hard and soft localization objectives and background-aware grading supervision.\\
\textbf{Results:} Evaluated on a private dataset of 1,859 LC videos, STC-Net achieves an accuracy of 62.11\% and an F1-score of 61.42\%, outperforming non-localized baselines by over 10\% in both metrics and highlighting the effectiveness of weak supervision for surgical complexity assessment.\\
\textbf{Conclusion:} STC-Net demonstrates a scalable and effective approach for automated PGS-based surgical complexity estimation from full LC videos, making it promising for post-operative analysis and surgical training.
}




\maketitle

\section{Introduction}
Laparoscopic cholecystectomy (LC) is one of the most common surgical procedures for the treatment of cholecystitis \cite{Csikesz2010}. Yet, its intraoperative course can vary significantly due to the wide spectrum of underlying pathologies, ranging from mild biliary colic to gangrenous cholecystitis \cite{Ward2022}. Severe inflammation is associated with greater technical difficulty, longer operative times, and increased risk of complications \cite{Lee2020}.

\begin{figure*}[t]
    \centering
    \includegraphics[width=1\textwidth]{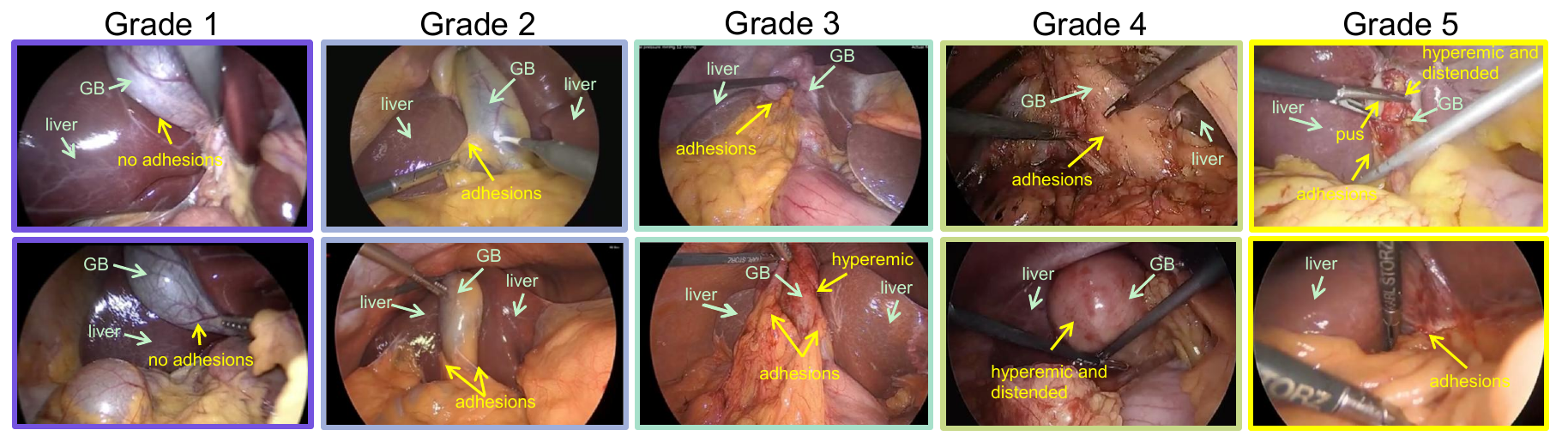}
    \caption{
    Examples of different PGS grades from our dataset. The screenshots correspond to the moment at which the PGS grade was assigned. GB stands for gallbladder. 
    } \label{fig_grade_examples}
   
\end{figure*}

Given the variability of gallbladder inflammation, reliable grading systems are essential for stratifying LC complexity. Such systems can inform intraoperative decision-making \cite{Sugrue2019} and enable case stratification to support surgical training and education. Among those, such as G10, Nassar, and the Parkland Grading Scale (PGS) \cite{Sugrue2019, Griffiths2019, Madni2018}, the PGS is widely adopted for its objective criteria and strong inter-rater reliability \cite{Madni2018, Ward2022}. It is a 5-point scale assigned early in the operation based on visual and anatomical features such as adhesions, distension, and necrosis \cite{Madni2018}. Higher PGS grades correlate with greater operative difficulty and the risk of complications, underscoring the need for an accurate assessment \cite{Madni2019}. Representative examples of PGS grades are shown in Fig.~\ref{fig_grade_examples} (see Suppl. Materials for detailed descriptions).

The growing availability of surgical video has enabled the automation of many applications in LC, such as workflow recognition, CVS detection, and anatomy segmentation \cite{Nwoye2022,Funke2025,Zhang2024,Hong2020,Mehta2024,Grammatikopoulou2024}. Building on these advances, automating PGS estimation offers a standardized alternative to subjective intraoperative assessment by surgeons, thus providing a reproducible measure of surgical complexity.

Automated PGS estimation aims to predict the grade of gallbladder inflammation severity from LC videos. This grade is typically assigned during the initial inspection phase of the gallbladder. To support model training, two types of annotations are required: (i) the PGS label and (ii) the temporal window (start and end timestamps) that defines the grading period. In real clinical settings, the model must estimate the PGS grade directly from raw, full-length videos without relying on manually trimmed clips. This requires the model to learn how to localize the relevant segment for grading during training. While providing full temporal annotations (\textit{i.e.,} start and end timestamps) can facilitate model training, such annotations are time-consuming to generate and often ambiguous. A more practical alternative is to annotate a single timestamp that marks a representative moment for grading. This form of weak temporal supervision can effectively guide the model in localizing informative segments for PGS estimation, offering a middle ground between fully supervised and fully unlocalized (no temporal annotations) approaches.

In this work, we propose \textbf{STC-Net}, a novel framework for \textbf{S}ingle-\textbf{T}imestamp-based \textbf{C}omplexity estimation in LC using the PGS. To the best of our knowledge, STC-Net is the first method to estimate PGS directly from full-length LC videos without manual trimming at test time, making it suitable for real postoperative use. Trained under weak temporal supervision \cite{Ma2020,Xia2024}, where each video is labeled with a single timestamp and its PGS grade, STC-Net jointly localizes informative segments and predicts the PGS grade. It comprises three components: a \textit{Localization Module} (LM) that assigns frame-wise relevance scores, a \textit{Window Proposal Module} (WPM) that extracts candidate temporal segments based on localization confidence, and a \textit{Grading Module} (GM) that classifies segments and aggregates PGS predictions via a consensus strategy. A novel supervision scheme further enforces synergy between localization and grading objectives. Our main contributions are:

\begin{itemize}
    \item We propose STC-Net, the first framework to estimate PGS in LC directly from full-length surgical videos.
    \item We design the WPM that dynamically generates informative temporal segments around predicted timestamps. This allows for video-specific adaptive slicing, improving the relevance of the inputs to the grading model.
    \item We introduce a novel training strategy that jointly optimizes localization and grading objectives. This includes a two-part localization loss that combines hard (binary) and soft (distributional) supervision, and a background-aware grading loss that improves robustness by explicitly modeling non-informative segments.
    \item We evaluate STC-Net through extensive experiments on a large private dataset of LC procedures.
\end{itemize}

\noindent\textbf{Related Work:}
Existing works that address PGS (or other grading scale) estimation in LC videos often simplify the task by framing it as frame-wise classification, training and evaluating models on isolated static images. Abbing \emph{et al.} \cite{Abbing2023} trained Convolutional Neural Networks (CNNs) on still images labeled with a modified Nassar score. Wu \emph{et al.} \cite{Wu2023} proposed a framework for surgical phase recognition, CVS detection and PGS estimation, and Ward \emph{et al.}~\cite{Ward2022} used CNNs to classify PGS grades from static images. While these approaches may perform well on curated images, they do not reflect the real-world scenario, where the specific frames relevant to grading are unknown at test time. Moreover, inflammation severity may not be evident in still frames, requiring temporal cues such as tissue manipulation or bleeding for accurate assessment. A more temporally oriented approach was proposed by Ban \emph{et al.}~\cite{Ban2024}, trained and evaluated on 16-second video clips capturing the initial inspection of the gallbladder. It uses a Concept Graph Neural Network to model surgical concepts associated with inflammation. However, despite leveraging temporal information, their setup still assumes prior knowledge of when grading occurs (testing on trimmed clips).

\section{Methods}

For a surgical video with $T$ frames, we pre-compute frame-level representations, resulting in features $X \in \mathbb{R}^{T\times D}$, where $D$ is the feature dimension. Each video is annotated with a PGS class label denoted by $c \in \{1, \dots, C\}$, where $C$ is the number of PGS classes, and a target timestamp $t \in \mathbb{N}$ marking the temporal location where the grade was assigned. The timestamp can be converted into a one-hot vector $y \in \{0,1\}^T$ as $y_i = 1$ if $i = t$, and $y_i = 0$ otherwise, for $i = 0, \dots, T-1$. This forms the basis of our weakly supervised setting, where the model is trained using $(X, y, c)$.

\subsection{Framework} \label{sec_framework}

\begin{figure*}[t]
    \centering
    \includegraphics[width=1.0\textwidth]{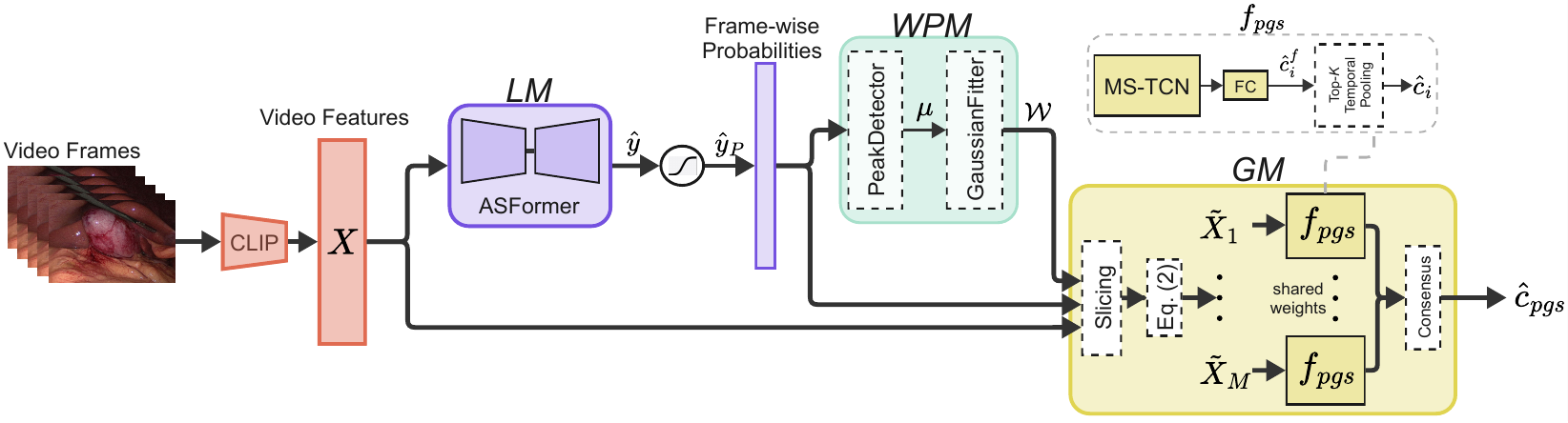}
    \caption{
Overview of STC-Net: A frozen CLIP encoder extracts frame features $X \in \mathbb{R}^{T \times D}$. The Localization Module (LM) predicts frame-wise probabilities $\hat{y}_P$, which the Window Proposal Module (WPM) converts into candidate windows $\mathcal{W}$. The Grading Module (GM) integrates $X$, $\hat{y}_P$, and $\mathcal{W}$ to predict the final grade $\hat{c}_{\text{pgs}}$.
    } \label{fig_framework}
   
\end{figure*}


As shown in Fig.~\ref{fig_framework}, given a surgical video encoded into features $X$, STC-Net first applies the LM to generate frame-wise scores indicating the likelihood of each frame corresponding to the target timestamp. These scores are then passed to the WPM, which proposes candidate temporal segments (start and end timestamps) most informative for the estimation of PGS. For each proposal, the corresponding feature slice of $X$ is extracted and processed independently by the GM, which aggregates the individual predictions to yield the final PGS estimate.

\noindent\textbf{Feature Extraction:}
To obtain frame-level visual representations $X$, we leverage the visual encoder of CLIP \cite{Radford2021}. It is pre-trained on a large-scale in-house dataset of LC videos using text prompts describing surgical phases and anatomical structures.

\noindent\textbf{Localization Module:}
The LM takes CLIP features $X \in \mathbb{R}^{T\times D}$ as input and outputs frame-level scores $\hat{y} \in \mathbb{R}^T$. We implement LM with ASFormer \cite{Yi2021}, which is well-suited for modeling long-range temporal dependencies. 
Frame-wise probabilities can be obtained by applying a sigmoid activation to $\hat{y}$, resulting in $\hat{y}_{p} \in \left[ 0,1 \right]^{T}$, where each value indicates the estimated probability that the corresponding frame matches the target timestamp. 
The predicted target timestamp is $\hat{t} = \arg\max_{i \in \{0, \dots, T-1\}} \hat{y}_i$.

\noindent\textbf{Window Proposal Module:}
The WPM aims to extract temporally localized segments around key frames predicted as informative for grading by the LM, serving as focused input for the downstream PGS classification task. To identify candidate temporal regions, we apply a peak detection function $\mathrm{PeakDetector}(\cdot)$ to $\hat{y}_P$, which identifies $K$ local maxima using a probability threshold. The resulting peak indices $\mu = [\mu_1, \dots, \mu_M]$ indicate frames with high confidence in being the target timestamp, and $M$ denotes the number of proposals. We fit two Gaussians on either side of each peak located at $\mu_i$, estimating the standard deviations $\sigma_{l,i}$ and $\sigma_{r,i}$ for the left and right portions of the signal $\hat{y}_P$, respectively. Formally, for each side, we fit an one-dimensional Gaussian function (via non-linear least squares) of the form:

\begin{equation} \label{eq_gaussian}
     g(\tau) = A \cdot \exp\left(-(\tau - \mu_i)^2/2\sigma_i^2  \right),
\end{equation}
where $\tau \in [0, \mu_i]$ for the left side and $\tau \in [\mu_i, T]$ for the right side of the peak, $A$ denotes the amplitude, and $\sigma_i \in \{\sigma_{l,i}, \sigma_{r,i}\}$. The standard deviations are used to dynamically define the window boundaries around the peak, controlled by a scale hyperameter $N_{std}$, as $(l_i, r_i) = \mu_i \pm N_{\mathrm{std}} \cdot (\sigma_{l,i}, \sigma_{r,i})$. The resulting set of proposals is denoted by $\mathcal{W} = \{(l_1, r_1), \dots, (l_M, r_M)\}$ (see Alg. 1 of Suppl. Materials). Unlike fixed-length windows, our dynamic approach adapts window size based on localization confidence.

\noindent\textbf{Grading Module:} The GM processes the set of candidate temporal windows identified by the WPM to make PGS predictions. For each window $(l_i, r_i)$ in $\mathcal{W}$, features $X_i$ and probabilities $\hat{y}_{Pi}$ are extracted via a slicing operation ($[l_i:r_i]$) and combined as:

\begin{equation} 
\tilde{X}_i = X_i \odot \hat{y}_{P_i} + X_i, \label{eq_product}
\end{equation}
where \(\odot\) denotes element-wise multiplication. This operation amplifies regions of high confidence while preserving the original signal with the skip connection. Additionally, it maintains a direct computational graph connection between the LM and the GM, enabling gradient flow from the classification objective back to the localization.

The reweighted feature segment $\tilde{X}_i$ is then forwarded to the PGS model $f_{pgs}(\cdot)$, which produces raw classification logits $\hat{c}_i \in \mathbb{R}^{C+1}$, where the additional class at index 0 denotes background (\emph{i.e.}, frames deemed uninformative for PGS estimation). This process is applied to all $M$ proposals, yielding a set of predictions $\{\hat{c}_1, \dots, \hat{c}_M\}$. A consensus block $\mathrm{Consensus}(\cdot)$ is then applied over this set to derive the final grade prediction $\hat{c}_{\text{pgs}} \in \{1, \dots, C\}$ (see Alg. 2 of Suppl. Materials). 

As shown in Fig.~\ref{fig_framework}, the PGS model $f_{pgs}(\cdot)$ uses MS-TCN \cite{Farha2019} for temporal modeling within each proposal. The reweighted segment $\tilde{X}_i$ is fed into the MS-TCN to produce frame-wise logits $\hat{c}_i^f \in \mathbb{R}^{T_i \times (C+1)}$, which are then aggregated by top-$K$ pooling \cite{Ma2020} into the proposal level prediction $\hat{c}_i \in \mathbb{R}^{C+1}$. As for the $\mathrm{Consensus}(\cdot)$ block, we select the proposal with the highest localization confidence $i^* = \arg\max_{i} \hat{y}[\mu_i]$, and predict the most confident (non-background) class, $\hat{c}_{\text{pgs}} = \arg\max_{j \in \{1, \dots, C\}} \hat{c}_{i^*, j}$.

\section{Experimental Design}
\subsection{Training Objectives}

\noindent\textbf{Localization:} We employ a dual-loss strategy composed of a custom Binary Cross-Entropy (BCE) loss and a cosine similarity loss. 

The BCE loss treats the localization as frame-wise binary classification, where the label vector $y \in \{0,1\}^T$ has $y_t = 1$ at the target timestamp and $0$ elsewhere. To reduce penalties on frames temporally close to the target, we define a neutral zone of size $3\delta$ around $t$. 
Negative loss terms are applied only to frames outside this zone, while $\delta \in \mathbb{N}^+$ controls the tolerance. The BCE loss is defined as:

\begin{equation}
\mathcal{L}_{\text{bce}} =  
- y_{t} \log \hat{y}_{P,t} 
- \frac{1}{T-1} \sum\limits_{\substack{j = 0 \\ j \notin [t - 3\delta,\, t + 3\delta]}}^{T - 1}
(1 - y_j) \log(1 - \hat{y}_{P,j})
\end{equation}

While BCE enforces sharp separation between the target and other frames, it ignores that nearby frames may contain relevant cues. To soften supervision, we introduce a cosine similarity loss that aligns the softmax-normalized predictions $\hat{y}_{\text{soft}}$ (along the temporal dimension of $\hat{y}$) with a Gaussian reference $\mathcal{N}(t,\delta^2) \in \mathbb{R}^T$ centered at $t$, and with standard deviation of $\delta$. We define this loss as:

\begin{equation}
\mathcal{L}_{\text{cos}} = 1 - \frac{\hat{y}_{\text{soft}} \cdot \mathcal{N}(t, \delta^2)}{\left\| \hat{y}_{\text{soft}} \right\|_2 \cdot \left\| \mathcal{N}(t, \delta^2) \right\|_2}
\end{equation}

The $\mathcal{L}_{\text{bce}}$ enforces precision at the target frame, while $\mathcal{L}_{\text{cos}}$ encourages smoother scores around it, 
mitigating overfitting from temporal ambiguity or annotation noise. The total localization loss is $\mathcal{L}_{\text{L}} = \mathcal{L}_{\text{bce}} + \alpha \mathcal{L}_{\text{cos}}$, where $\alpha$ is a weight hyperparameter.

\noindent\textbf{Grading:} The grading loss is cross-entropy over all $M$ proposals: the proposal whose peak $\mu_i$ is closest to $t$ 
(\emph{i.e.,} $i^+ = \arg\min_{i} \left|\mu_i - t\right|$) is treated as the positive sample with video label $c$, while all other proposals 
are negatives assigned to the background class (0). The final loss combines cross-entropy on the positive proposal 
with a penalty encouraging negatives to be classified as background:

\begin{equation} \label{grading_loss_eq}
\mathcal{L}_{\text{G}} =
-\log\left( 
\frac{\exp(\hat{c}_{i^+, c})}{\sum_{j=0}^{C} \exp(\hat{c}_{i^+, j})} 
\right)
- \frac{1}{M-1} \sum_{\substack{i=1 \\ i \neq i^+}}^{M} 
\log\left( 
\frac{\exp(\hat{c}_{i,0})}{\sum_{j=0}^{C} \exp(\hat{c}_{i, j})} 
\right) 
\end{equation}

\noindent\textbf{Training Scheme:} We adopt a two-stage training scheme. In the first stage, only the LM is trained using the localization objective $\mathcal{L}_{\text{L}}$, while the GM remains frozen. This allows the LM to identify informative timestamps without interference from the grading task. After a fixed number of epochs, marked by the frozen epoch $E_{frozen}$, we switch to joint training, with the total loss defined as $\mathcal{L} = \mathcal{L}_G + \beta \mathcal{L}_L$, where $\beta$ controls the contribution of the localization loss.

\subsection{Dataset and Evaluation}

We evaluate STC-Net on a private dataset of 1,859 LC videos, each annotated with a PGS grade (1–5) and a single timestamp by trained annotators. The class distribution is: Grade 1 (443), Grade 2 (318), Grade 3 (444), Grade 4 (445), and Grade 5 (209). A held-out test set of 190 videos is used for evaluation. For PGS classification, we use Accuracy, Precision, Recall, F1-Score, and Average Distance (AD), defined as the mean absolute error between ground-truth and predicted grades. For localization, we use the Mean Absolute Error (MAE) between predicted and ground-truth timestamps.

\subsection{Implementation Details}
STC-Net is trained with Adam for 100 epochs ($lr=1 \times 10^{-5}$, batch size 1, $E_{\text{frozen}}=8$). Videos are downsampled to 1Hz, and $D$ is 768. The ASFormer uses an encoder and a 3-stage decoder (10 layers each, feature size 64). The MS-TCN uses 1 stage with 2 layers, feature size 64, dropout 0.2. $K=8$ for top-$K$ pooling. Peaks are detected using SciPy's \texttt{find\_peaks} ($\text{threshold}=0.5$), and Gaussians are fitted with SciPy's \texttt{curve\_fit}. Other parameters are set as $\delta=50$, $\alpha=\beta=1$, and $N_{std}=2$. STC-Net is implemented in PyTorch v2.4 and trained on an NVIDIA RTX A6000 (48GB).

\section{Results and Discussion}

\subsection{Comparison with Baselines} \label{section_baselines_results}
To the best of our knowledge, no prior work has addressed PGS estimation from full-length surgical videos that jointly perform temporal localization and grading. Therefore, to contextualize the performance of our approach, we design a set of representative baselines for comparison. The \textit{Full} baseline processes the entire video as input, without any use of the target timestamp during training or test-time. \textit{Full} serves as a lower bound, highlighting the difficulty of learning from long videos where irrelevant content can dilute relevant information for grading. To approximate an upper bound with ideal localization, we use the \textit{Trimmed} baseline, which uses ground-truth timestamps at test time. Videos are cropped to symmetric windows around the annotated timestamp for training and testing (see Table~\ref{table_main}). For a fair comparison, all baselines use the same precomputed CLIP features and GM architecture as STC-Net.

As shown in Table~\ref{table_main}, STC-Net significantly outperforms the \textit{Full} baseline, with +10.83\% and +10\% increases in accuracy and F1-Score, respectively, highlighting the importance of temporal localization for downstream PGS estimation, even when provided only through weak supervision. Also, STC-Net performs on par with the \textit{Trimmed} variant (20s), demonstrating that its weakly supervised training alone can match the performance of a model with ground-truth localization access at test time.

\begin{figure*}[t]
    \centering
    \includegraphics[width=1\textwidth]{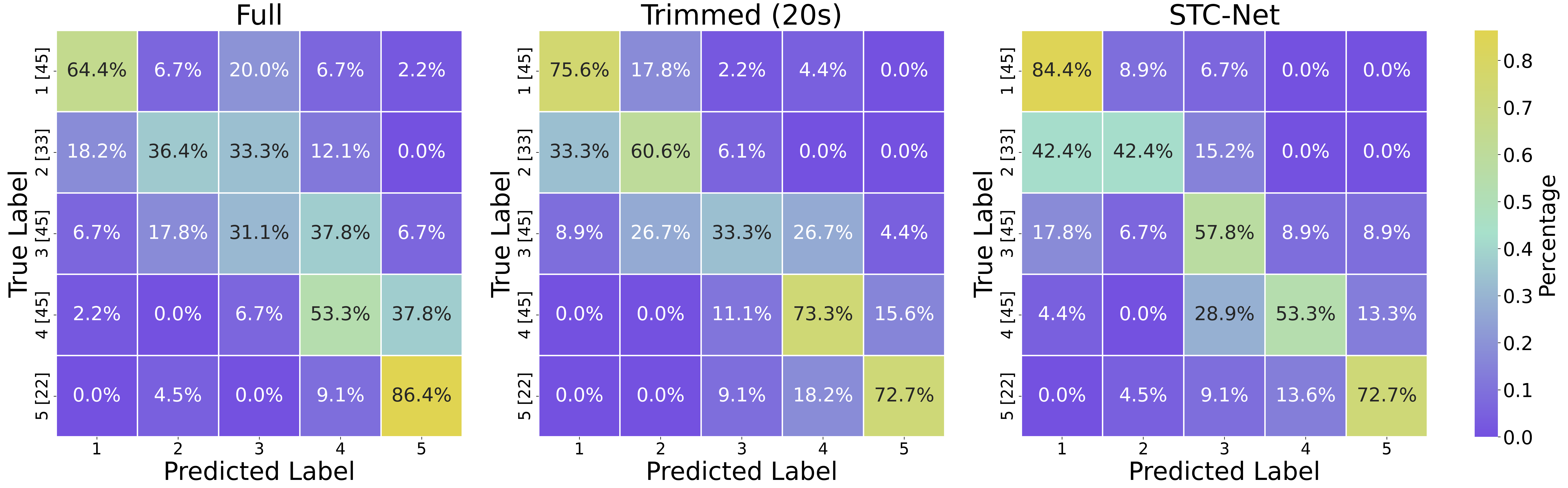}
    \caption{Confusion matrices for \textit{Full}, \textit{Trimmed (20s)}, and STC-Net (left to right) on the PGS classification task. Numbers in $[.]$ indicate the number of videos per class.} \label{fig_confmat}
\end{figure*}

To analyze model behavior, we present confusion matrices in Fig.~\ref{fig_confmat}. STC-Net performs well on Grades 1 and 5, while showing some confusion among intermediate grades, particularly Grades 2 and 4. However, most misclassifications occur between adjacent classes, indicating errors are generally mild. This aligns with the low AD of 0.5, showing incorrect predictions remain close to the true grade. This is important in clinical practice as the model rarely confuses extreme grades, thus minimizing the risk of severely under- or overestimating disease severity. The \textit{Trimmed} baseline (20s) handles Grades 2 and 4 slightly better, likely due to its access to ground-truth timestamps at test time. In contrast, STC-Net performs better on Grade 1 and nearly doubles accuracy on Grade 3. These results show that minimal, low-cost timestamp supervision can match methods relying on manually trimmed (ideal) localization.

Furthermore, a reason why all models tend to underperform on the intermediate classes may stem from the clinical definition of the PGS itself, where overlapping criteria \cite{Madni2018} (\textit{e.g.,} presence of adhesions) can blur the distinction between adjacent grades in terms of inflammation severity, whereas Grades 1 and 5 exhibit more distinct visual characteristics, making them easier to discriminate reliably. Finally, while STC-Net and the \textit{Trimmed} baseline outperform the \textit{Full} model on most classes, the latter does better on Grade 5. This suggests that for accurate Grade 5 prediction, localization, whether precise (\textit{Trimmed}) or estimated (STC-Net), offers no benefit; instead, access to the entire video is required. Clinically, Grade 5 reflects severe, widespread inflammation that may manifest in multiple surgical phases, so global video context may be needed to capture its defining features. For example, the gallbladder may be obscured by adhesions, with additional signs of inflammation only visible after their removal.

\begin{table*}[t]
\centering
\caption{Baseline performance on the PGS classification task. *\underline{Without} and \textsuperscript{\textdagger}\underline{with} use of target timestamps at train/test time.}

\label{table_main}
\resizebox{1.0\textwidth}{!}{%
\begin{tabular}{c|c|c|cccccc}
\hline
\textbf{Model Input}           & \textbf{Loc.  Supervision}                         & \textbf{Method} & \textbf{Accuracy ($\uparrow$)} & \textbf{Precision ($\uparrow$)} & \textbf{Recall ($\uparrow$)} & \textbf{F1-Score ($\uparrow$)} & \textbf{AD ($\downarrow$)} & \textbf{MAE ($\downarrow$)} \\ \hline
Full Videos                    & None                                                       & *Full           & 51.58                          & 51.54                           & 54.32                        & 51.42                          & 0.65                       & NA                          \\ \hline
\multirow{5}{*}{\begin{tabular}[c]{@{}c@{}}Trimmed \\ Clips\end{tabular}} & \multirow{5}{*}{\begin{tabular}[c]{@{}c@{}}NA \\ (manually clipped, \\ ideal localization)\end{tabular}}
 & \textsuperscript{\textdagger}Trimmed (20s)   & \textbf{62.11}                 & 61.62                           & \textbf{63.11}               & 61.37                          & \textbf{0.45}              & \multirow{5}{*}{NA}         \\
                               &                                                            & \textsuperscript{\textdagger}Trimmed (60s)   & 61.57                          & 61.48                           & 62.83                        & 61.37                          & 0.48                       &                             \\
                               &                                                            & \textsuperscript{\textdagger}Trimmed (120s)  & 57.37                          & 57.32                           & 57.39                        & 57.23                          & 0.51                       &                             \\
                               &                                                            & \textsuperscript{\textdagger}Trimmed (180s)  & 61.05                          & 61.09                           & 60.48                        & 60.30                          & 0.48                       &                             \\
                               &                                                            & \textsuperscript{\textdagger}Trimmed (240s)  & 59.47                          & 59.62                           & 58.85                        & 58.84                          & 0.52                       &                             \\ \hline
Full Videos                    & single timestamp                          & \textbf{STC-Net}         & \textbf{62.11}                 & \textbf{63.39}                  & 62.14                        & \textbf{61.42}                 & 0.50                       & \textbf{89.94}              \\ \hline
\end{tabular}%
}
\end{table*}

\subsection{Ablation Studies}

\begin{table}[t]
\centering
\caption{Ablation studies of the Window Proposal Module.}\label{table_wpm}%
\begin{tabular}{cc|l|ccc}
\hline
\textbf{WPM}            & \textbf{GaussianFitter} & \multicolumn{1}{c|}{\textbf{Settings}} & \textbf{Accuracy} ($\uparrow$) & \textbf{F1-Score} ($\uparrow$) & \textbf{AD} ($\downarrow$) \\ \hline
\ding{55}                  & \ding{55}                  & \multicolumn{1}{c|}{entire sequence}   & 47.89             & 47.61             & 0.65                       \\ \hline
\multirow{5}{*}{\ding{51}} & \multirow{5}{*}{\ding{55}} & window of 20s                          & 57.37             & 54.56             & 0.55                       \\
                        &                         & window of 60s                          & 60.00             & 58.48             & 0.52                       \\
                        &                         & window of 120s                         & 53.16             & 49.82             & 0.63                       \\
                        &                         & window of 180s                         & 58.42             & 57.40             & 0.54                       \\
                        &                         & window of 240s                         & 60.00             & 58.97             & 0.51                       \\
\multicolumn{1}{l}{}    & \multicolumn{1}{l|}{}   & window of 300s                         & 55.26             & 53.23             & 0.59                       \\ \hline
\ding{51}                  & \ding{51}                  & \multicolumn{1}{c|}{STC-Net}           & \textbf{62.11}    & \textbf{61.42}    & \textbf{0.50}              \\ \hline
\end{tabular}%
\end{table}

\noindent\textbf{Analysis of the Window Proposal Module:}
We assess key design choices in the WPM through experiments with the following configurations: (i) We remove the WPM entirely. Without proposal generation, the entire sequence is fed directly into the GM. Eq.~\ref{eq_product} reduces to $\tilde{X} = X \odot \hat{y}_{P} + X $
and no slicing is applied. This experiment evaluates the benefit of explicitly focusing on localized segments versus processing the full video; (ii) We replace the dynamic window generation algorithm ($\mathrm{GaussianFitter}$) with predefined fixed-size windows centered around each detected peak. This configuration isolates the impact of dynamic window sizes tailored to each video. Table~\ref{table_wpm} presents the results for ablations (i) and (ii). 

Ablation (i) highlights the critical role of the WPM: removing it and feeding the full sequence into the GM leads to a -14.22\% accuracy drop. This setup even underperforms the \textit{Full} baseline (Table~\ref{table_main}), which does not use any localization, suggesting that localization and window-level slicing should be combined to be effective. Ablation (ii) shows that dynamic window proposals with fitted Gaussians outperform fixed-size (240s) slicing by +2.11\% in accuracy, indicating that variable-length windows are better suited to capturing video-specific temporal context. This gain likely stems from different videos containing varying numbers of frames that are informative for PGS estimation, something dynamic windows adapt to more effectively than fixed-size ones. Ablation (ii) also shows that even fixed-size slicing around detected peaks outperforms models without slicing (first row of Table~\ref{table_main} and Table \ref{table_wpm}), highlighting the importance of providing the GM with focused temporal proposals (\textit{i.e.,} slices) rather than full sequences (except for Grade 5).

\noindent\textbf{Analysis of Training Losses and Scheme:} \label{section_ablation_losses}
Table~\ref{table_losses_and_scheme} shows the contribution of each localization loss ($\mathcal{L}_{\text{bce}}$, $\mathcal{L}_{\text{cos}}$) and the background term of the grading loss (2nd term of Eq.~\ref{grading_loss_eq}, denoted as $\mathcal{L}_{\text{bg}}$). Removing $\mathcal{L}_{\text{bg}}$ corresponds to supervising only the positive window, without explicit background modeling.

\begin{figure*}[t]
    \centering
    \includegraphics[width=1\textwidth]{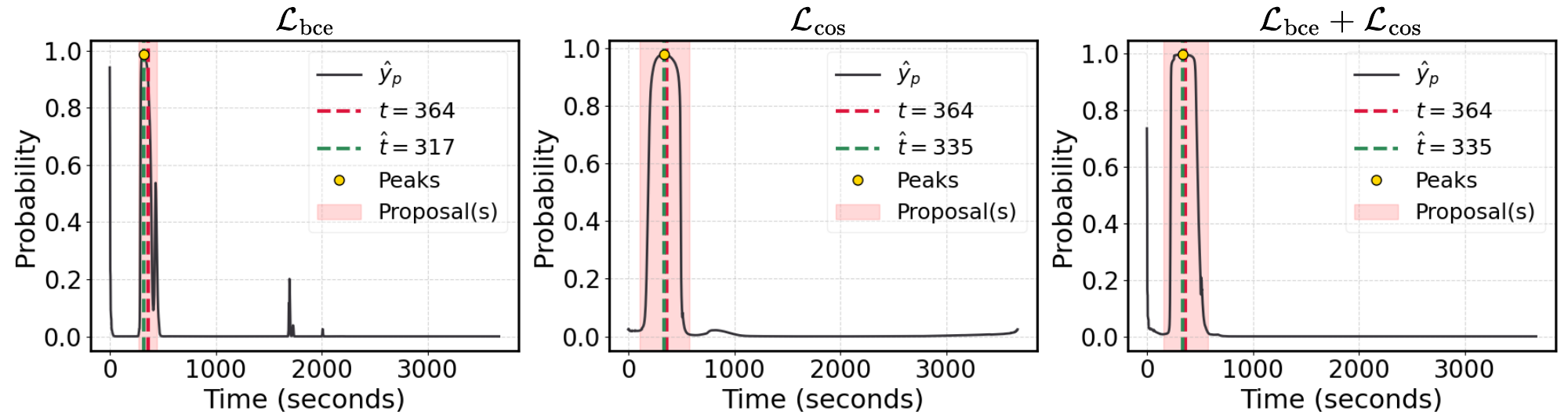}
    \caption{Qualitative comparison of frame-wise localization probabilities $\hat{y}_P$ under different supervision: from left to right, $\mathcal{L}_{\text{bce}}$, $\mathcal{L}_{\text{cos}}$, and $\mathcal{L}_{\text{bce}}+\mathcal{L}_{\text{cos}}$.} \label{fig_loc_output_ablation_losses}

\end{figure*}

The combination of $\mathcal{L}_{\text{bce}}$ and $\mathcal{L}_{\text{cos}}$ outperforms either loss alone. We attribute this to $\mathcal{L}_{\text{bce}}$ enforcing sharp peaks at the target timestamp without penalizing neutral frames, while $\mathcal{L}_{\text{cos}}$ provides soft supervision by encouraging a smooth Gaussian-like shape around it, thereby handling uncertainty. Interestingly, using only $\mathcal{L}_{\text{cos}}$ outperforms $\mathcal{L}_{\text{bce}}$, suggesting distribution-based supervision is more robust to annotation noise and temporal ambiguity. As shown in Fig.~\ref{fig_loc_output_ablation_losses}, $\mathcal{L}_{\text{bce}}$ yields sharp peaks which can lead to proposals that miss important context, while $\mathcal{L}_{\text{cos}}$ yields smoother, wider peaks that may include redundant frames. Their combination balances these effects, producing proposals better suited for grading.

\begin{table}[t]
\centering
\caption{Ablating losses and training schemes.}
\label{table_losses_and_scheme}
\begin{tabular}{ccc|cccc}
\hline
$\mathcal{L}_{bce}$                                                                              & $\mathcal{L}_{cos}$ & $\mathcal{L}_{bg}$ & Accuracy ($\uparrow$) & F1-Score ($\uparrow$) & AD ($\downarrow$) & MAE ($\downarrow$) \\ \hline
\ding{51}                                                                                           & \ding{55}              & \ding{51}             & 56.84                 & 54.90                 & 0.55              & 98.35              \\
\ding{55}                                                                                           & \ding{51}              & \ding{51}             & 58.95                 & 57.39                 & 0.54              & 94.44              \\
\ding{51}                                                                                           & \ding{51}              & \ding{55}             & 57.89                 & 57.50                 & 0.54              & \textbf{81.45}     \\
\ding{51}                                                                                           & \ding{51}              & \ding{51}             & \textbf{62.11}        & \textbf{61.42}        & \textbf{0.50}     & 89.94              \\ \hline
\multicolumn{1}{c|}{\multirow{3}{*}{\begin{tabular}[c]{@{}c@{}}\textbf{Training} \\ \textbf{Scheme}\end{tabular}}} & \multicolumn{2}{c|}{End-to-End}          & 57.37                 & 56.87                 & 0.56              & 130.57             \\
\multicolumn{1}{c|}{}                                                                            & \multicolumn{2}{c|}{Separate}            & 61.05                 & 60.58                 & \textbf{0.49}     & 100.11             \\
\multicolumn{1}{c|}{}                                                                            & \multicolumn{2}{c|}{Two-Stage}           & \textbf{62.11}        & \textbf{61.42}        & 0.50              & \textbf{89.94}     \\ \hline
\end{tabular}%
\end{table}

Removing the $\mathcal{L}_{\text{bg}}$ term from the grading loss results in a noticeable drop of -4.22\% in grading accuracy, showing the value of explicitly modeling uninformative frames as background, which helps the model better differentiate between relevant and irrelevant temporal content. However, localization MAE improves in the absence of $\mathcal{L}_{\text{bg}}$, but this is expected as the optimization becomes more focused on the localization objectives.

\begin{table}[h]
\centering
\caption{Impact of different consensus types.}
\label{table_consensus_type}%
\begin{tabular}{l|ccc}
\hline
\textbf{Consensus} & \textbf{Accuracy} ($\uparrow$) & \textbf{F1-Score} ($\uparrow$) & \textbf{AD} ($\downarrow$) \\ \hline
Average            & 59.47             & 58.81             & 0.53              \\
Majority Vote      & 61.05             & 60.60             & 0.52              \\
Highest Confidence & 61.58             & 60.86             & 0.51              \\
Highest Peak (proposed)       & \textbf{62.11}             & \textbf{61.42}             & \textbf{0.50}              \\ \hline
\end{tabular}%
\end{table}

We further study the impact of the training scheme (see Table~\ref{table_losses_and_scheme}). We compare three strategies: \textit{End-to-End} training, where LM and GM are jointly optimized from the start; \textit{Separate} training, where the LM is trained first and then frozen while training the GM; and our proposed \textit{Two-Stage}. Our \textit{Two-Stage} scheme achieves the best overall performance by first letting the LM focus on localization and then refining it with grading supervision. \textit{End-to-End} underperforms likely because early joint optimization yields weak proposals that hurt grading. \textit{Separate} performs better than \textit{End-to-End}, but still worse than \textit{Two-Stage} since the LM is never updated with grading loss.

\noindent\textbf{Impact of Consensus Block:}\label{section_consensus_ablation} 
We further compare four proposal prediction aggregation strategies for the $\mathrm{Consensus}(\cdot)$ block: \textit{Average} (mean score across proposals), \textit{Majority Vote} (most frequent class), \textit{Highest Confidence} (class from proposal with highest class confidence), and \textit{Highest Peak} (see Section~\ref{sec_framework}). The background class is always excluded. As shown in Table~\ref{table_consensus_type}, naive averaging yields the weakest performance, as confident predictions may be diluted by noisy ones. Although class confidence-based strategies \textit{Majority Vote} and \textit{Highest Confidence} yield stronger results, focusing on the most confident proposal (\textit{Highest Peak}) proves even more effective.

\section{Conclusion}
We presented STC-Net, the first framework to estimate surgical complexity via the PGS from full-length LC videos, trained using weak temporal supervision. STC-Net consists of a LM for timestamp prediction, a novel WPM that generates candidate segments, and a GM that classifies them and aggregates their predictions. The framework is trained with a novel combination of soft and hard temporal supervision to align localization and grading objectives, and a background-aware loss to suppress uninformative frames. Extensive experiments on a large-scale dataset demonstrate strong performance, highlighting STC-Net’s effectiveness and its potential for real-world post-operative use. Future work will focus on real-time adaptation, and incorporating semantic priors and multi-scale temporal modeling to improve grading performance.

\backmatter

\bmhead{Supplementary information}

This article has accompanying supplementary files. 

\bmhead{Acknowledgements}
This work was supported by Medtronic plc.; the EPSRC under the UCL Doctoral Training Partnership (DTP) [EP/R513143/1, EP/T517793/1], the UCL Centre for Doctoral Training in Intelligent, Integrated Imaging in Healthcare (i4health) [EP/S021930/1] and the Human-centred Machine Intelligence to optimise Robotic Surgical Training (HuMIRoS) project [EP/Z534754/1]; the Department of Science, Innovation and Technology (DSIT) and the Royal Academy of Engineering under the Chair in Emerging Technologies programme; For the purpose of open access, the author has applied a CC BY public copyright licence to any author accepted manuscript version arising from this submission. This work was done during D. Anastasiou's internship at Medtronic plc.

\section*{Declarations}
\bmhead{Conflict of interest}
S. Barbarisi, L. Culshaw, J. Patel, and I. Luengo are employees of Medtronic plc. D. Stoyanov is an employee of Medtronic plc and UCL.
\bmhead{Ethical approval}
Medtronic plc maintains all necessary rights and consents to process, analyze and display the private data referenced in this study. The work is conducted through redacted videos obtained through commercial or research agreements, and without Protected Health Information (PHI).

\bibliographystyle{bst/sn-aps}
\bibliography{ipcai}

@article{Yi2021,
  title={ASFormer: Transformer for Action Segmentation},
  author={Fangqiu Yi and Hongyu Wen and Tingting Jiang},
  journal={BMVC},
  year={2021},
}

@inproceedings{Radford2021,
  title={Learning transferable visual models from natural language supervision},
  author={Radford, Alec and Kim, Jong Wook and Hallacy, Chris and others},
  booktitle={ICML},
  pages={8748--8763},
  year={2021}
}

@article{Farha2019,
   author = {Yazan Abu Farha and Jurgen Gall},
   issn = {10636919},
   journal = {CVPR},
   title = {MS-TCN: Multi-stage temporal convolutional network for action segmentation},
   pages = {3575-3584},
   year = {2019},
}

@article{Ma2020,
  title     = {SF-Net: Single-Frame Supervision for Temporal Action Localization},
  author = {Ma, Fan and Zhu, Linchao and Yang, Yi and others},
  journal = {ECCV},
  year      = {2020},
  pages     = {415--431},
}

@article{Ban2024,
  author    = {Yutong Ban and Jennifer A. Eckhoff and others},
  title     = {Concept Graph Neural Networks for Surgical Video Understanding},
  journal   = {IEEE Trans Med Imaging},
  year      = {2024},
  volume    = {43},
  number    = {1},
  pages     = {264--274},
}

@article{Madni2018,
title = {The Parkland grading scale for cholecystitis},
journal = {Am J Surg},
volume = {215},
number = {4},
pages = {625-630},
year = {2018},
issn = {0002-9610},
author = {Tarik D. Madni and David E. Leshikar and Christian T. Minshall and others},
}

@article{Csikesz2010,
  author    = {Nicholas G. Csikesz and Anand Singla and Melissa M. Murphy and others},
  title     = {Surgeon Volume Metrics in Laparoscopic Cholecystectomy},
  journal   = {Dig Dis Sci},
  year      = {2010},
  volume    = {55},
  number    = {8},
  pages     = {2398--2405},
  month     = August,
  issn      = {1573-2568}
}

@article{Ward2022,
  author    = {Thomas M. Ward and Daniel A. Hashimoto and Yutong Ban and others},
  title     = {Artificial intelligence prediction of cholecystectomy operative course from automated identification of gallbladder inflammation},
  journal   = {Surg Endosc},
  year      = {2022},
  volume    = {36},
  number    = {9},
  pages     = {6832--6840},
  month     = September,
  issn      = {1432-2218}
}

@article{Lee2020,
title = {Does surgical difficulty relate to severity of acute cholecystitis? Validation of the parkland grading scale based on intraoperative findings},
journal = {Am J Surg},
volume = {219},
number = {4},
pages = {637-641},
year = {2020},
issn = {0002-9610},
author = {Woohyung Lee and Jae Yool Jang and Jin-Kyu Cho and others}
}

@article{Sugrue2019,
  author    = {Michael Sugrue and Federico Coccolini and Maciej Bucholc and others},
  title     = {Intra-operative gallbladder scoring predicts conversion of laparoscopic to open cholecystectomy: a WSES prospective collaborative study},
  journal   = {World J Emerg Surg},
  year      = {2019},
  volume    = {14},
  pages     = {12},
  month     = mar,
  issn      = {1749-7922},
  pmid      = {30911325},
  pmcid     = {PMC6417130}
}

@article{Griffiths2019,
  author    = {E.A. Griffiths and J. Hodson and others},
  title     = {Utilisation of an operative difficulty grading scale for laparoscopic cholecystectomy},
  journal   = {Surg Endosc},
  year      = {2019},
  volume    = {33},
  number    = {1},
  month     = jan,
  issn      = {1432-2218},
  pmid      = {29956029},
  pmcid     = {PMC6336748},
}

@article{Madni2019,
title = {Prospective validation of the Parkland Grading Scale for Cholecystitis},
journal = {Am J Surg},
volume = {217},
number = {1},
pages = {90-97},
year = {2019},
issn = {0002-9610},
author = {Tarik D. Madni and Paul A. Nakonezny and others}
}

@article{Nwoye2022,
title = {Rendezvous: Attention mechanisms for the recognition of surgical action triplets in endoscopic videos},
journal = {MedIA},
volume = {78},
pages = {102433},
year = {2022},
issn = {1361-8415},
author = {Chinedu Innocent Nwoye and Tong Yu and others}
}

@article{Hong2020,
  author       = {Wen-Yu Hong and Chih-Lun Kao and Yu-Huan Kuo and others},
  title        = {CholecSeg8k: A Semantic Segmentation Dataset for Laparoscopic Cholecystectomy Based on Cholec80},
  year         = {2020},
  eprint       = {2012.12872},
  archivePrefix= {arXiv}
}

@article{Mehta2024,
  author    = {Pritesh Mehta and David Owen and Maria Grammatikopoulou and others},
  title     = {Hierarchical segmentation of surgical scenes in laparoscopy},
  journal   = {IJCARS},
  year      = {2024},
  volume    = {19},
  number    = {7},
  pages     = {1449--1457},
  month     = July,
  issn      = {1861-6429}
}

@article{Grammatikopoulou2024,
  author    = {Maria Grammatikopoulou and Ricardo Sanchez-Matilla and Felix Bragman and others},
  title     = {A spatio-temporal network for video semantic segmentation in surgical videos},
  journal   = {IJCARS},
  year      = {2024},
  volume    = {19},
  number    = {2},
  pages     = {375--382},
  month     = February,
  issn      = {1861-6429}
}

@article{Funke2025,
  author={Funke, Isabel and Rivoir, Dominik and others},
  journal={IEEE Trans. Biomed. Eng.}, 
  title={TUNeS: A Temporal U-Net With Self-Attention for Video-Based Surgical Phase Recognition}, 
  year={2025},
  volume={},
  number={},
  pages={1-15},
}

@article{Zhang2024,
  author    = {Jinglu Zhang and Santiago Barbarisi and Abdolrahim Kadkhodamohammadi and others},
  title     = {Self-knowledge distillation for surgical phase recognition},
  journal   = {IJCARS},
  year      = {2024},
  volume    = {19},
  number    = {1},
  month     = jan,
  issn      = {1861-6429}
}

@article{Abbing2023,
author = {Julian. R. Abbing and Frank J. Voskens and others},
title = {Towards an AI-based assessment model of surgical difficulty during early phase laparoscopic cholecystectomy},
journal = {Comput. Methods Biomech. Biomed. Eng.},
volume = {11},
number = {4},
pages = {1299--1306},
year = {2023},
publisher = {Taylor \& Francis},
}

@article{Wu2023,
  author    = {Shangdi Wu and Zixin Chen and Runwen Liu and others},
  title     = {SurgSmart: an artificial intelligent system for quality control in laparoscopic cholecystectomy: an observational study},
  journal   = {Int J Surg},
  year      = {2023},
  volume    = {109},
  number    = {5},
  pages     = {1105--1114},
  month     = may,
}

@InProceedings{Xia2024,
    author    = {Xia, Ziying and Cheng, Jian and Liu, Siyu and others},
    title     = {Realigning Confidence with Temporal Saliency Information for Point-Level Weakly-Supervised Temporal Action Localization},
    booktitle = {CVPR},
    month     = {June},
    year      = {2024},
    pages     = {18440-18450}
}

\end{document}